\documentclass{article}

\usepackage{xcolor} 
\usepackage{graphicx} 
\graphicspath{ {./images/} } 
\usepackage{arxiv}
\usepackage{kotex}
\usepackage[utf8]{inputenc} 
\usepackage[T1]{fontenc}    
\usepackage{hyperref}       
\usepackage{url}            
\usepackage{booktabs}       
\usepackage{amsfonts}       
\usepackage{nicefrac}       
\usepackage{microtype}      
\usepackage{lipsum}
\usepackage{multirow}
\usepackage{subcaption}
\title{Real-time Mask Detection on Google Edge TPU}

\author{
  Keondo Park, Wonyoung Jang, Woochul Lee, Kisung Nam, Kihong Seong, Kyuwook Chai, Wen-Syan Li \\[4mm]
  Department of Data Science\\[1mm]
  Seoul National University, Korea\\[2mm]
  \texttt{\{gundo0102, jwy4888, woochulee, kisung.nam, kev94yo, manapool, wensyanli\}@snu.ac.kr}
}

\begin{document}
\maketitle

\begin{abstract}
After the COVID-19 outbreak, it has become important to automatically detect whether people are wearing masks in order to reduce risk of front-line workers.
In addition, processing user data locally is a great way to address both privacy and network bandwidth issues.
In this paper, we present a light-weighted model for detecting whether people in a particular area wear masks, which can also be deployed on Coral Dev Board, a commercially available development board containing Google Edge TPU.
Our approach combines the object detecting network based on MobileNetV2 plus SSD and the quantization scheme for integer-only hardware.
As a result, the lighter model in the Edge TPU has a significantly lower latency which is more appropriate for real-time execution while maintaining accuracy comparable to a floating point device.
Code is available at: \href{https://github.com/KeondoPark/coral\_mask}{\texttt{https://github.com/KeondoPark/coral\_mask}}
\end{abstract}


\section{Introduction}


After the COVID-19 pandemic, the necessity of wearing masks in public places cannot be overemphasized.
To prevent the spread of the virus, monitoring whether people are wearing masks has become an important task.
However, it is quite inefficient to manually check who is wearing a mask or not. Especially, front-line workers face a great risk because they stay in places with a large floating population and checking whether all people wear masks are burdensome.
In addition, there are many public places that are susceptible to virus infection because many people can be crowded in a particular area.
Therefore, installing cameras for monitoring in all of those places could entail a large cost.
To mention another problem, if the camera simply records a video and transmits it to the cloud for the inference, privacy and network bandwidth issues can arise.
In order to address these problems, processing data locally, especially on low-cost devices, can be a great solution.


In this paper, we present a light-weighted model for detecting whether people are wearing masks that could fit in Coral Dev Board \cite{coral-datasheet}, a commercially available development board containing Google Edge TPU coprocessor—a 8-bit, fixed point hardware.
Our approach combines the object detection network utilizing MobileNetV2 \cite{DBLP:journals/corr/abs-1801-04381} plus SSD \cite{10.1007/978-3-319-46448-0_2} and the quantization scheme \cite{10.5555/3122009.3242044, Jacob_2018_CVPR} for integer-only devices.
MobileNetV2 is a model for performing computer vision applications in a mobile devices. It features reduced computation using few parameters while retaining similar accuracy to complex CNN models.
Single Shot Detector (SSD) model detects objects from images with a single neural network. SSD predicts bounding boxes from feature maps from different resolutions, thereby handles different objects well. Our model combined MobileNetV2 and SSD, a single neural network.
We employed transfer learning from existing face detection neural network as well as integer-only post-training quantization.

As a result, we demonstrate that the lighter model in the Edge TPU has a significantly lower latency while maintaining accuracy in comparison to a floating point hardware.
This makes it plausible to detect whether people are wearing masks in real-time without privacy and network issues, consequentially widely applicable in the real world.

\section{Related Work}

Convolutional Neural Networks (CNN) has been widely used for image classification since AlexNet \cite{NIPS2012_4824}, and it has developed through VGG \cite{simonyan2014deep}, GoogLeNet \cite{7298594}, and ResNet \cite{7780459}.
However, these studies were focused on improving the accuracy of the model which increased the number of layers and parameters gradually without consideration of computational efficiency.
As artificial intelligence expands its application, deep neural network models are implemented on smartphones or edge computational devices. This encouraged many approaches to reduce latency and model size, while maintaining its accuracy.
In order to reduce computational complexity, SqueezeNet \cite{i2016squeezenet} applied 1$\times$1 convolution and reduced filter size and MobileNet \cite{howard2017mobilenets} employed depthwise-separable convolution.
MobileNetV2 \cite{DBLP:journals/corr/abs-1801-04381} achieved state-of-the-art results using inverted residuals and linear bottlenecks.
Another approach to improve computational efficiency of CNN is quantization \cite{10.5555/3122009.3242044, Jacob_2018_CVPR} that reduces the precision of the weight parameter rather than the number of parameters. Using quantization, the size of the model can be significantly reduced while maintaining accuracy.

Object detection methods can be categorized in two frameworks: region proposal networks such as R-CNN \cite{6909475}, Fast R-CNN \cite{girshick2015fast}, and Faster R-CNN \cite{10.5555/2969239.2969250} and regression/classification based networks such as MultiBox \cite{6909673}, YOLO \cite{7780460}, and SSD \cite{10.1007/978-3-319-46448-0_2}.
Region proposal networks make proposals for the potential region where the object lies, and then classify the object.
Regression/classification based networks consider object detection and classification tasks as a single regression task.
Since regression/classification based networks process the input image directly to bounding boxes and probabilities, it takes less time compared to region proposal networks.
In this paper, we utilized the architecture of SSD \cite{10.1007/978-3-319-46448-0_2} to ensure the accuracy and performance in edge computational devices.

\section{Model} 
In this section, we describe our model.



\subsection{Architecture}
We started our journey with two neural network structure, namely 2NN model. This model uses two different neural networks: one for detection and the next one for classification. We used MobileNetV2 plus SSD model to detect faces and another MobileNetV2 model to classify the detected face into mask or no mask. We did not train the detection model, but used the pre-trained model from tensorflow model zoo\cite{tfmodel_zoo}. We trained the classification model with our data by transfer learning from MobileNetV2.

In order to improve the model efficiency we incorporated the classification model into detection model, namely 1NN model. 1NN model uses MobileNetV2 plus SSD architecture\cite{DBLP:journals/corr/abs-1801-04381} \cite{10.1007/978-3-319-46448-0_2}. As will be described in the following section, our dataset has relatively small number of images. Therefore we took advantage of transfer learning from pre-trained face detection model. We took the face detection model from Tensorflow 1 Object Detection Model Zoo\cite{tfmodel_zoo} and used this model as our base model. Before we start training, we made some change on base model, for better transfer learning: Since the face detection model has only one class(face) but our model needed two classes(Mask, NoMask), we extended the class prediction layer to have 2 classes. We copied the weights corresponding to face detection from the base model and applied same weights for both mask and nomask class in our model.  For mask class, however, we slightly changed the weights from the base model: we added 1e-7 to each weight corresponding to mask class. By doing so, the initial model would not give the same score for both mask and nomask class but give priority to the nomask class. We did nit change weights for image feature maps and box prediction. After this setup is completed, we started our transfer learning - this is explained in the following section.

\begin{figure}[h!]
    \centering
    \begin{subfigure}[b]{1\linewidth}
      \includegraphics[width=\linewidth, height=6cm]{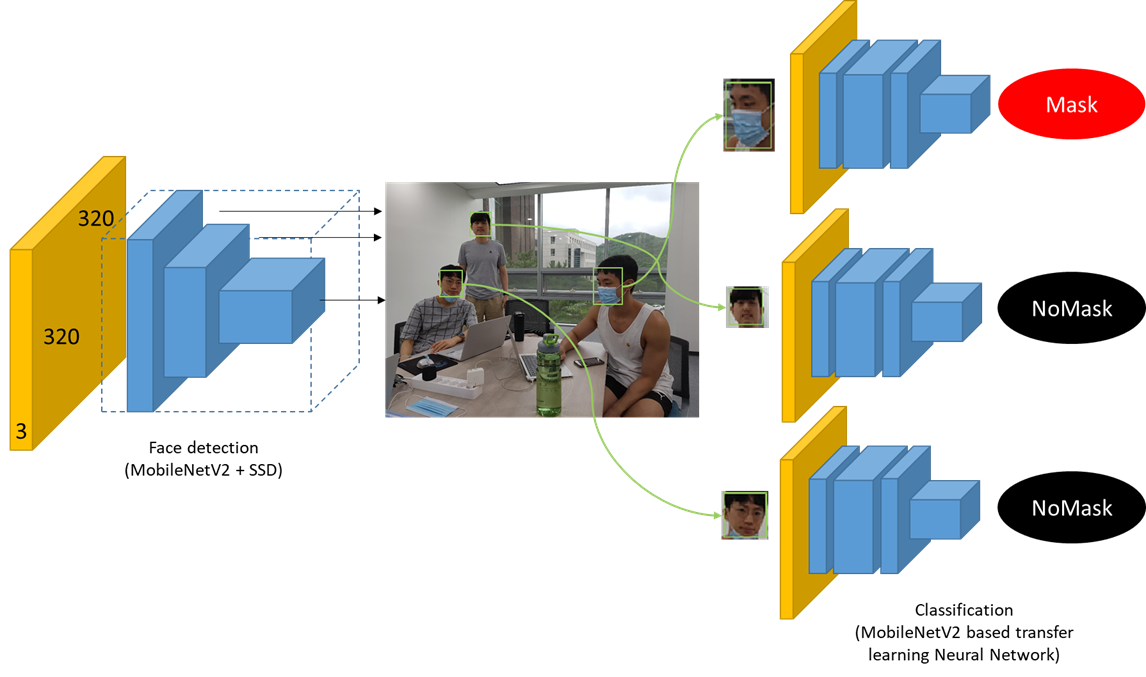}
      \caption{2NN model architecture}
    \end{subfigure}
    \begin{subfigure}[b]{1\linewidth}
      \includegraphics[width=\linewidth, height=6cm]{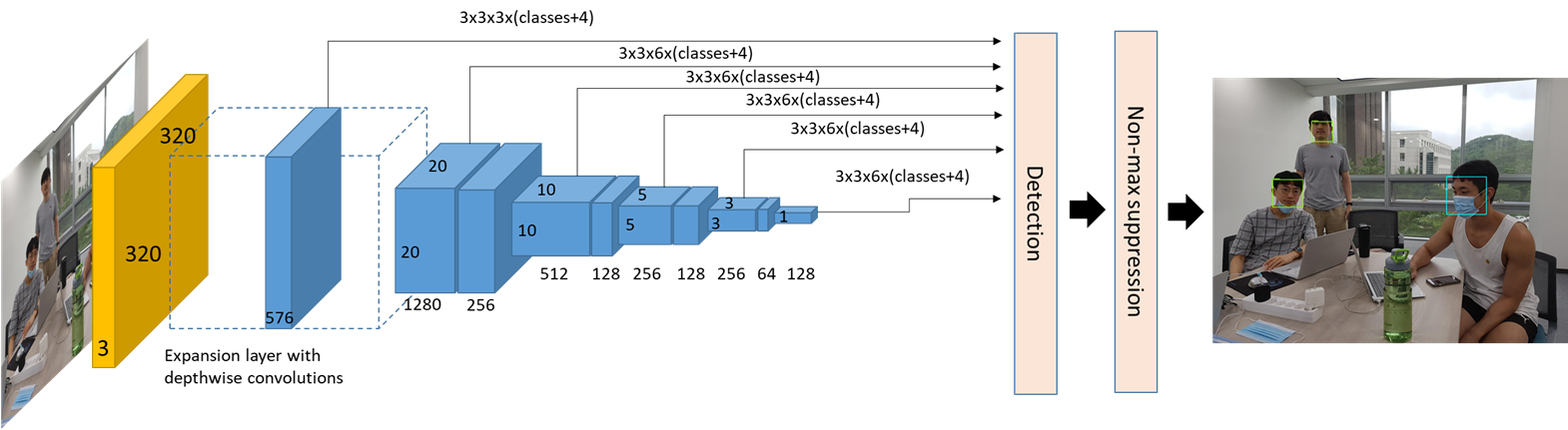}
      \caption{1NN model architecture}
    \end{subfigure}
    \caption{Comparison of 1NN and 2NN model architecture}
    \label{fig:modelArchitecture}
\end{figure}

The high level comparison of 1NN and 2NN model architecture is described in Figure \ref{fig:modelArchitecture}

The second neural network in 2NN model is comprised of MobileNet v2 for classification + Average Pooling, Flatten and Dense Layers.
The second neural network has additional layer on top of MibileNetV2. All the parameters in the base MobileNetV@ layer is set 'non-trainable'. Next, two-dimensional (2D) average pooling layer is added for down-sampling of tensor. And flatten layer is applied followed by dense layer. The output of dense layer is (None, 128) with activation function of 'RELU'. And then, drop out layer is applied for regularization fraction of the input units with dropout ratio of 0.5. The training scheme is described in 

\begin{figure}[h] 
\centering
\includegraphics[width=0.5\textwidth]{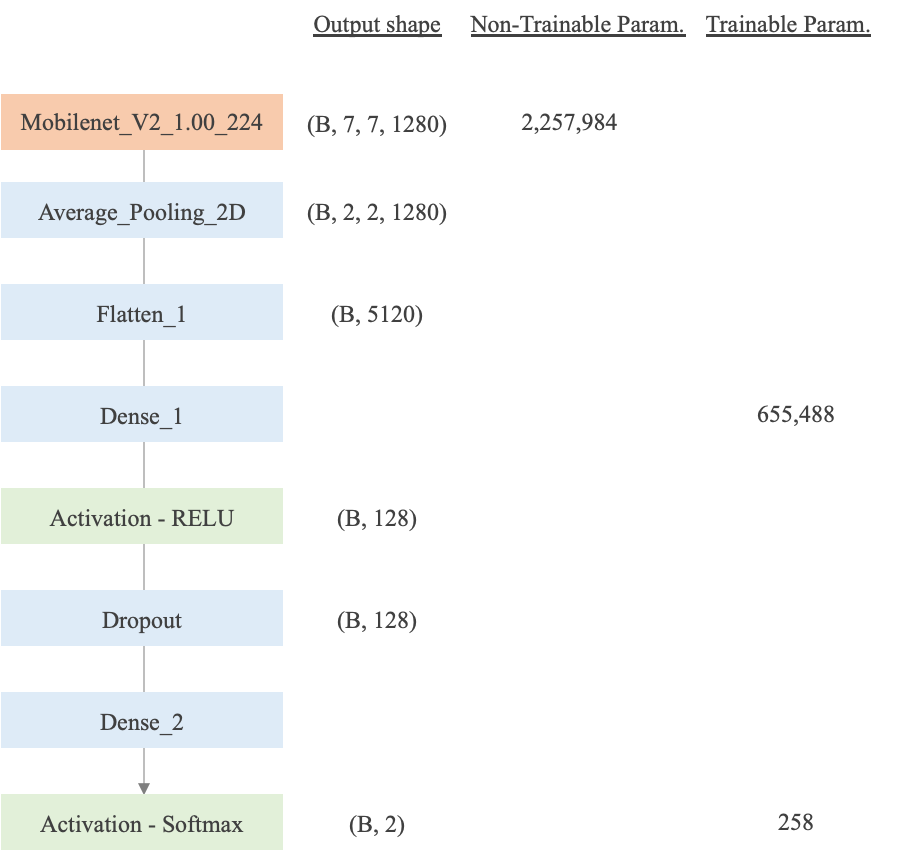}
\caption{Architecture of the second network in 2NN}
\end{figure}

\subsection{Data}
\begin{figure}[h!]
    \centering
    \begin{subfigure}[b]{0.35\linewidth}
      \includegraphics[width=\linewidth, height=6cm]{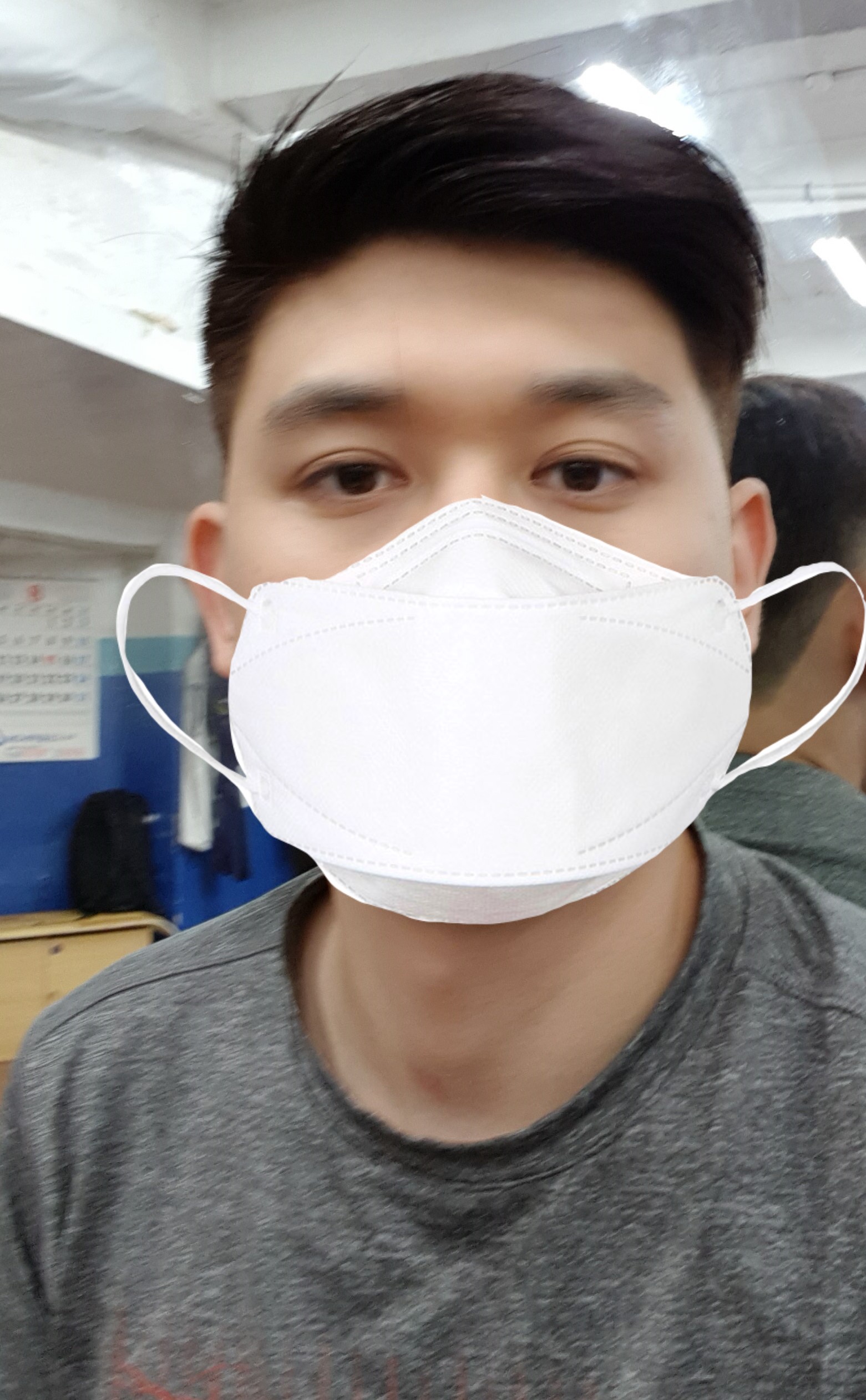}
      \caption{Artificial image from non-mask image}
    \end{subfigure}
    \begin{subfigure}[b]{0.35\linewidth}
      \includegraphics[width=\linewidth, height=6cm, angle=90]{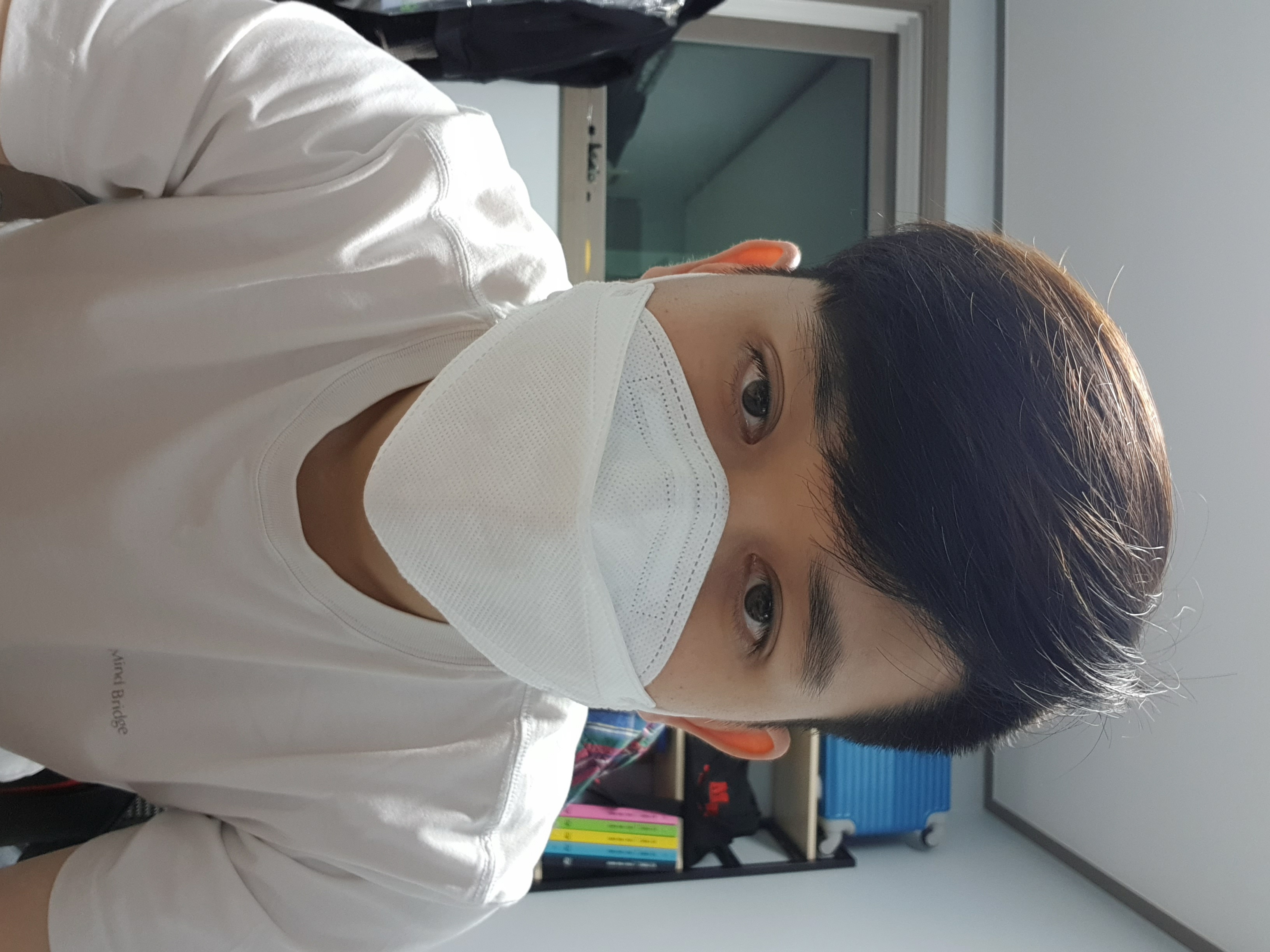}
      \caption{Real mask image}
    \end{subfigure}
    \caption{Example of mask images used for training}
    \label{fig:artificialMask}
\end{figure}

We collected 2,127 images of people's faces wearing masks and not wearing masks. We split the dataset into training set of 1,270 images, validation set of 425 images and test set of 425 images. When splitting the dataset, we took special care so that images with people wearing masks or not should be evenly distributed across training, validation and test set. Since the face images with masks on were fewer than raw face images, we added mask on top of faces, as in Figure \ref{fig:artificialMask}(a). This helps to balance the number of data between mask and nomask class.


\subsubsection{1NN Model}
For 1NN model, we marked the groundtruth boxes and labels in all faces in each image The number of images for training, validation and test set is same as described above.

\subsubsection{2NN Model}
For 2NN model, we did not train the first network for face detection as mentioned in the previous section. The second network is trained to distinguish whether the detected face is with a mask or without a mask. Therefore we cropped faces from training and validation set, and used these faces for training and validation set of 2NN model, respectively. Because some images we collected contain more than one face, there are greater number of data for 2NN model training and evaluation. The resulting number of images are 1,893 for training/validation. We used the same evaluation set as the 1NN model in order to consistently compare the performance of 1NN and 2NN models.

\subsection{Training}

\subsubsection{1NN Model}
\begin{figure}[h!]
    \centering
    \begin{subfigure}[b]{0.4\linewidth}
      \includegraphics[width=\linewidth]{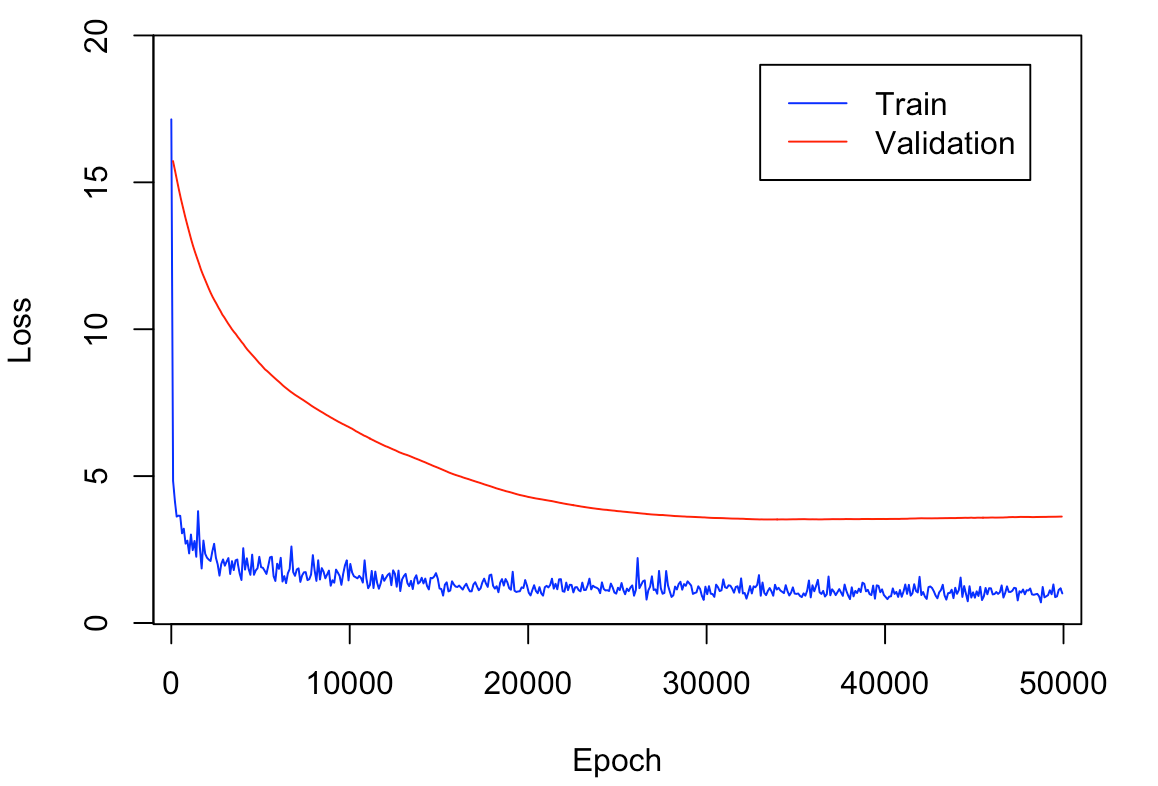}
      \caption{Base model}
    \end{subfigure}
    \begin{subfigure}[b]{0.4\linewidth}
      \includegraphics[width=\linewidth]{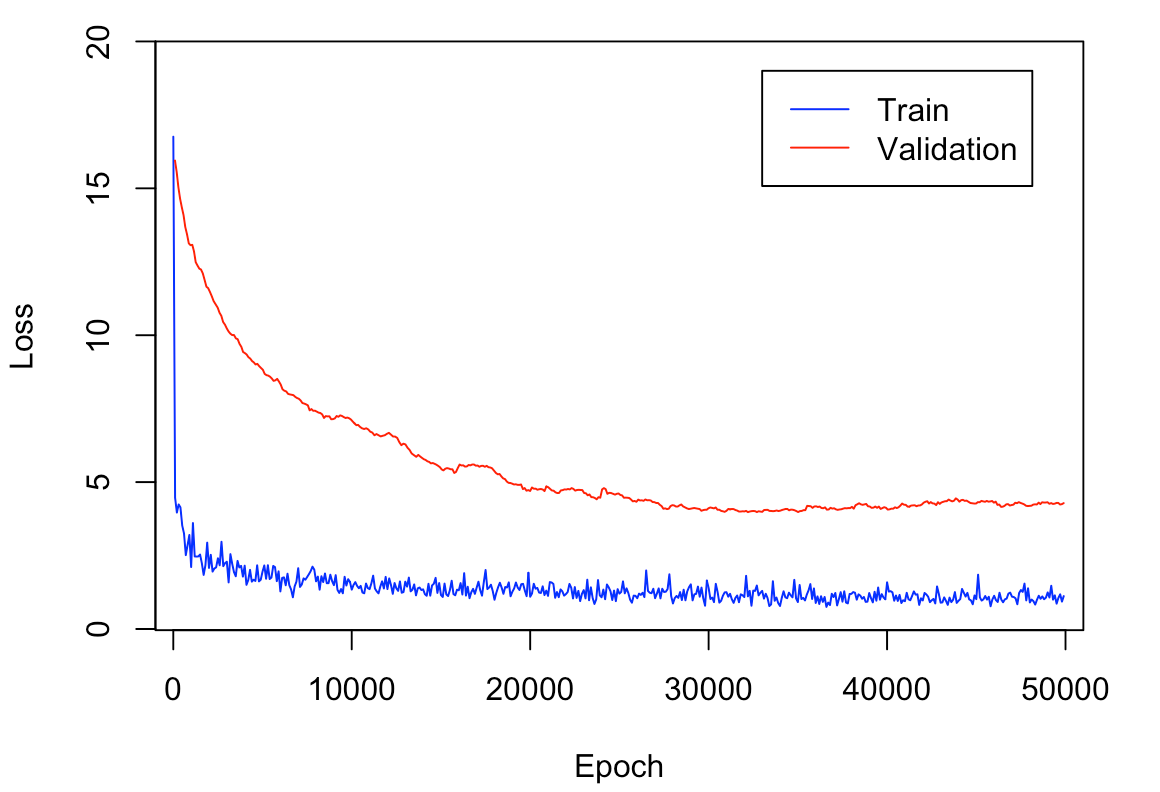}
      \caption{Quantized model}
    \end{subfigure}
    \caption{Train and validation loss of the model}
    \label{fig:Loss}
\end{figure}

In transfer learning process, we did not freeze any layers in the base MobilenetV2 plus SSD model. We used batch size of 32. We used RMS Prop optimizer with initial learning rate of 0.04, momentum optimizer value of 0.9 and decaying at 0.9. The learning rate is also subject to exponential decaying schedule with decaying factor of 0.95 and decaying step of 800,720. Images are pre-processed to have 320 X 320 resolution, before used for training. We did quantization-aware training for quantized model and same hyperparameters were used for both models. Both models were trained for 50,000 epochs. The training and validation losses are described in \ref{fig:Loss}. We used Intel Xeon Silver 4216 CPU and Nvidia Titan RTX GPU 1EA for training.


\subsubsection{2NN Model}
\paragraph{Training Setup}
We train our models using TensorFlow\cite{tensorflow2015-whitepaper}. We used GPU server. GPU server is built by Intel Xeon Silver 4216 CPU(32 cores, 2.1GHz), 512GB memory and Nvidia Titan RTX GPU 1EA. Initial learning rate is 1e-2 and learning rate decay rate is initial learning rate divided by the number of each epoch. Image size is 224 X 224. Input images are pre-processed to 224 X 224, whatever magnitudes they have. Binary cross entropy was used as loss function. The number of epochs is 32 and batch size is 32.

\section{Experimental Evaluation}
In this section, we describe our evaluation methodology and results.

\subsection{Model Size}
1NN model has 5.5 million parameters and its size is 22.9MB before quantization. The model size decreased to 5.4MB after 8-bit full integer quantization. The number of parameters increased slightly because of the additional parameters required for quantization. Special type of compiling is necessary to run the model on EdgeTPU. This so called EdgeTPU compiling increases the size of model to 6.4MB.

The first detection model in 2NN model has 3.4 million parameters and its size is 22.2MB before quantization and 5.2MB after quantization. The second classification model has 2.9 million parameters and its size is 17.2MB before quantization and 3.5MB after quantization. After EdgeTPU compiling, the model size increases to 6.4MB and 3.55MB, respectively.

\begin{table}[h!]
  \centering
  \begin{tabular}{lllll}
    \toprule
    \multirow{2}[2]{*}{Model} &\multirow{2}[2]{*}{Params} &\multicolumn{3}{c}{Size}                   \\
    \cmidrule{3-5}
                       &       & Baseline        & Quantization  &EdgeTPU compiling \\
    \midrule
    1NN                & 5.5M  & 22.9MB &5.4MB  & 6.4MB  \\
    2NN Detection      & 3.4M     & 22.2MB      &5.2MB & 6.4MB  \\
    2NN Classification & 2.9M  & 17.2MB      &3.5MB & 3.55MB  \\
    \bottomrule
  \end{tabular}
  \vspace{2mm}
  \caption{Summary of model sizes}
  \label{tab:table}
\end{table}

 
\subsection{Experiment Environment}
For comparison, experiments of measuring latency and accuracy of 1NN and 2NN models are held in a CPU and an Edge TPU environment. Experiments for the CPU environment are conducted on Intel® Core™ i7-6700(3.40 GHz). Experiments for the Edge TPU environment are measured on Google Edge TPU coprocessor, included in the Coral Dev Board.

The Coral Dev Board is a single-board computer released by Google in 2019. The Coral Dev Board comprises an on-board System-on-Module(SoM). The SoM provides a fully-integrated system, especially including the Google Edge TPU coprocessor. The Google Edge TPU coprocessor is an ASIC designed by Google to perform machine learning inference on edge devices. The Google Edge TPU coprocessor is capable of performing 4 trillion operations (tera-operations) per second (TOPS), using 0.5 watts for each TOPS (2 TOPS per watt)\cite{coral-datasheet}.

For both environments, the 2NN Model is experimented only with the version after quantization. The 1NN model is experimented with both before and after quantization in both environments.

\subsection{Latency}
The amount of time to run a single inference is referred as latency, and is measured in this part. We measured the time spent for the inference of 425 test images and averaged the time spent for 1 image. Also we ran each model 3 times and averaged them. For the 2NN model, we need to go through two networks as mentioned before. Therefore, its latency is calculated as a sum of average inference time of face detection, and average inference time of mask classification. The 1NN model only utilizes one network, and the average inference time of that network is calculated as its latency. Three different versions are tested for each model; the baseline model, model after quantization, and model after Edge TPU compiling.  The experiment is done in both CPU and Edge TPU.

Overall, the 1NN model showed faster results, with the quantized version achieving 455ms of average inference time in the CPU environment, while the 2NN model took 1267ms for inference. In the Edge TPU environment, the 1NN model was still superior and showed 6.4ms of average inference time, while it took 26ms for the 2NN model. The baseline version of the 1NN model took 507ms for inference, which is faster than the quantized version of the 2NN model in a CPU environment. Latency testing results are summarized in Table \ref{tab:latency}.

\begin{table}[h!]
  \centering
  \begin{tabular}{llll}
    \toprule
    Environment & Data Type & 1NN & 2NN \\
    \midrule
    \multirow{2}{*}{CPU} & 32-bit & 507ms & - \\
        & 8-bit & 455ms & 1267ms \\
    \midrule
    Edge TPU & 8-bit & 6.4ms & 26ms \\
    \bottomrule
  \end{tabular}
  \vspace{2mm}
  \caption{Latency of the model according to different environments and architectures}
  \label{tab:latency}
\end{table}

\subsection{Accuracy}
The test accuracy is measured based on COCO mAP, which is widely used for evaluating image detection and classification results. Specifically, we measured AP at IoU=.50:.05:.95, which is used as the primary challenge metric. In the case of the 2NN model, mAP is not directly applicable, our evaluation needs few more steps. First, we get the bounding boxes and corresponding scores from the first model. Then we passed each bounding box through the second model and multiplied the score from the first model by that from the second model. This produces the same result set as if the test image passed through a single detection model, so we could apply the same evaluation metric. A total of 425 images are used as the test set, including images with more than 2 people. The test set also includes augmented images that combined masks on the bare face, and images of people wearing masks of various colors such as white, blue, black, and gray.

In conclusion, the difference in test accuracy according to the device was not large. First, for the 2NN model in 8-bit floating devices, the test accuracy of the CPU model was 55.0\%, and that of the Edge TPU model was 54.8\%. That is, there is little sacrifice in accuracy for Edge TPU compared to CPU. The same was true for the 1NN model. The test accuracy in 8-bit floating devices was 58.8\% for CPU model and 58.4\% for Edge TPU model, which was not a big difference. Compared to the 32-bit floating CPU, there was a difference of about 6\% points. Overall, the 1NN model showed better test accuracy than the 2NN model. Table \ref{tab:accuracy} summarizes the test accuracy.

\begin{table}[h!]
  \centering
  \begin{tabular}{llll}
    \toprule
    Environment & Data Type & 1NN & 2NN \\
    \midrule
    \multirow{2}{*}{CPU} & 32-bit & 64.7\% & - \\
        & 8-bit & 58.8\% & 55.0\% \\
    \midrule
    Edge TPU & 8-bit & 58.4\% & 54.8\% \\
    \bottomrule
  \end{tabular}
  \vspace{2mm}
  \caption{Accuracy of the model according to different environments and architectures}
  \label{tab:accuracy}
\end{table}

\section{Conclusion}

We presented a light-weighted convolutional neural network model for mask detection which is applicable in the real world in terms of latency and model size, by using MobileNetV2 plus SSD and quantization scheme.
Our model, even in a much smaller size compared to a model for 32-bit floating point devices, shows extremely low latency without sacrifice in accuracy.
This model can effectively avoid cost and network issues, accordingly can be widely used for public purposes in real life.








\bibliographystyle{unsrt}  
\bibliography{references}  

\end{document}